\documentclass{article}

\usepackage[preprint]{corl_2025} 
\usepackage{graphicx}

\newcommand{\name}{ExoStart}
\newcommand{\namespace}{ExoStart\space}

\makeatletter
\def\thickhline{%
  \noalign{\ifnum0=`}\fi\hrule \@height \thickarrayrulewidth \futurelet
   \reserved@a\@xthickhline}
\def\@xthickhline{\ifx\reserved@a\thickhline
               \vskip\doublerulesep
               \vskip-\thickarrayrulewidth
             \fi
      \ifnum0=`{\fi}}
\makeatother

\newlength{\thickarrayrulewidth}
\usepackage{array}

\title{\name:\\ Efficient learning for dexterous manipulation with sensorized exoskeleton demonstrations}

\author{Zilin Si \thanks{Authors with equal contribution.} $^{ ,}$ $^{1, 2}$, Jose Enrique Chen$^{*, 2}$, M. Emre Karagozler$^{*, 2}$, Antonia Bronars$^{3}$,  Jonathan Hutchinson$^{2}$, \\ \textbf{Thomas Lampe$^{2}$, Nimrod Gileadi$^{2}$, Taylor Howell$^{2}$, Stefano Saliceti$^{2}$, Lukasz Barczyk$^{2}$, } \\ 
\textbf{Ilan Olivarez Correa$^{2}$, Tom Erez$^{2}$, Mohit Shridhar$^{2}$, Murilo Fernandes Martins$^{2}$, } \\
\textbf{Konstantinos Bousmalis$^{2}$, Nicolas Heess$^{2}$, Francesco Nori$^{2}$, Maria Bauza Villalonga$^{2}$} \\
$^{1}$ Carnegie Mellon University, $^{2}$ Google DeepMind, $^{3}$ Massachusetts Institute of Technology\\
}

\begin{document}
\maketitle


\begin{abstract}
Recent advancements in teleoperation systems have enabled high-quality data collection for robotic manipulators, showing impressive results in learning manipulation at scale. This progress suggests that extending these capabilities to robotic hands could unlock an even broader range of manipulation skills, especially if we could achieve the same level of dexterity that human hands exhibit. However, teleoperating robotic hands is far from a solved problem, as it presents a significant challenge due to the high degrees of freedom of robotic hands and the complex dynamics occurring during contact-rich settings. In this work, we present ExoStart, a general and scalable learning framework that leverages human dexterity to improve robotic hand control. In particular, we obtain high-quality data by collecting direct demonstrations without a robot in the loop using a sensorized low-cost wearable exoskeleton, capturing the rich behaviors that humans can demonstrate with their own hands. We also propose a simulation-based dynamics filter that generates dynamically feasible trajectories from the collected demonstrations and use the generated trajectories to bootstrap an auto-curriculum reinforcement learning method that relies only on simple sparse rewards. The ExoStart pipeline is generalizable and yields robust policies that transfer zero-shot to the real robot. Our results demonstrate that ExoStart can generate dexterous real-world hand skills, achieving a success rate above 50\% on a wide range of complex tasks such as opening an AirPods case or inserting and turning a key in a lock. More details and videos can be found in \href{https://sites.google.com/view/exostart}{https://sites.google.com/view/exostart}.
\end{abstract}



\section{Introduction}
\label{sec:intro}
\begin{figure}[t]
\centering
\includegraphics[width=0.99\linewidth]{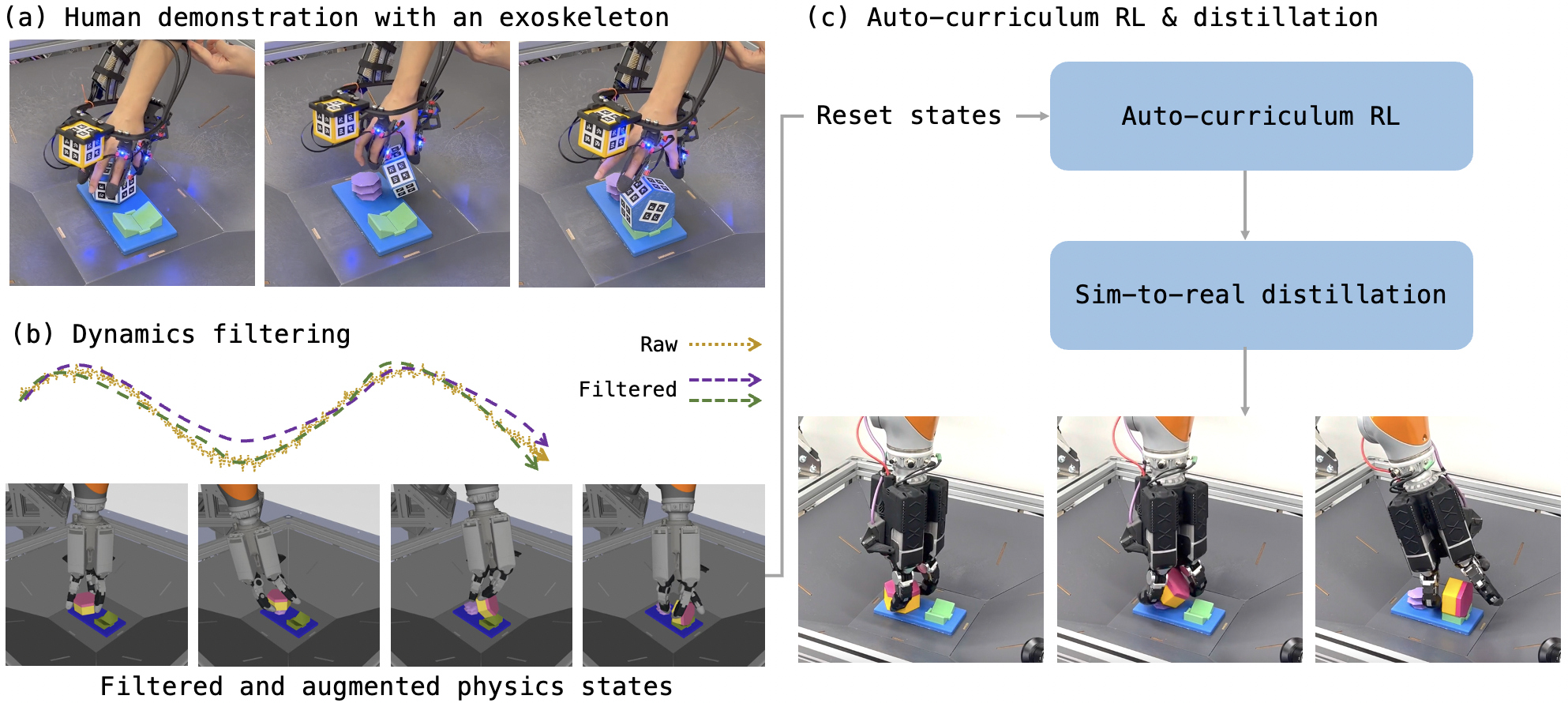}
\caption{Overview of the \namespace framework. \textbf{(a) Human demonstration with an exoskeleton}: We collect demonstrations with a sensorized exoskeleton by directly interacting with the object in the real world; \textbf{(b) Dynamics filtering}: We apply trajectory optimization to recover dynamically feasible simulated trajectories from the raw demonstrations; \textbf{(c) Auto-curriculum RL and distillation}: We use an auto-curriculum RL method~\cite{bauza2024demostart} to train a teacher policy from the filtered trajectories, and then distill it into a vision-based student policy that transfers zero-shot to the real world.}
\label{fig:overview}
\end{figure}

While robotic hands aim to replicate human dexterity, achieving versatile and dexterous manipulation still depends critically on  developing reliable and generalizable control methods. Recent advances on learning-based approaches for control have demonstrated promising results by successfully learning from high-quality data collected on systems with simple manipulators~\cite{team2025gemini, black2024pi_0}. However, extending these strategies to robotic hands remains challenging as it requires access to robust and scalable data collection systems, particularly when collecting high-quality data under contact-rich dynamics.

Despite recent progress in developing data collection systems, current methods remain limited and far from fully unlocking the dexterous potential of robotic hands. A common strategy involves teleoperating the robotic hands, often using wearable gloves~\cite{liu2019high, schwarz2021nimbro, wang2024dexcap, yin2025dexteritygen}, VR devices~\cite{arunachalam2023holo, cheng2024open, ding2024bunny}, or vision-based human hand tracking~\cite{handa2020dexpilot, sivakumar2022robotic, qin2023anyteleop, shaw2023leap}. However, these approaches often struggle with kinematic and dynamic mismatches between human and robotic hands, necessitating complex and specialized motion re-targeting, which often slows down and limits the expressiveness of the demonstrations. To address this, some approaches use a kinematic replica of the robotic hand for teleoperation~\cite{yang2024ace, si2024tilde, zhong2025nuexo}, which eliminates the kinematic gap and avoids motion re-targeting. Yet, these systems still rely on physical robots for data collection, leading to scalability challenges in terms of maintenance, complexity, and cost. Moreover, the inherent limitations of teleoperation, such as restricted haptic feedback, visibility, and latency, further hinder the demonstration of fine, dexterous, and reactive manipulation.

This work introduces a low-cost and scalable approach for collecting real-world demonstrations that capture the nuances of human dexterity without relying on a physical robot. Our solution, \name, is based on developing a sensorized wearable exoskeleton that is kinematically equivalent to a robotic hand. We design this exoskeleton as a glove, which enables operators to directly manipulate objects and feel the physical interactions as they demonstrate a task, as shown in Fig.~\ref{fig:overview}~(a). Embedded sensors on each joint record the finger motions of the exoskeleton, ensuring that the movements can be directly mapped to the robotic hand and thus avoiding the motion re-targeting problem. To mitigate noise and inconsistencies inherent in the real-world sensor data, we develop a dynamics filter~\cite{yamane2003dynamics} to recover trajectories that respect the simulated environment's dynamics. We then use the resultant trajectories to bootstrap an auto-curriculum RL method~\cite{bauza2024demostart} based on sparse rewards, to efficiently learn policies that can transfer zero-shot to the real robot.

Our proposed modular pipeline, \namespace (Fig.~\ref{fig:overview}), provides a reliable and robust recipe for collecting high-quality demonstrations and learning policies for real-world autonomous dexterous manipulation with robotic hands. Specifically, our main contributions include: (1) a scalable data collection setup based-on a low-cost sensorized exoskeleton for capturing expressive real-world human demonstrations without requiring a robot system in the loop; (2) a real-to-sim-to-real learning pipeline using dynamics filtering and auto-curriculum RL to efficiently learn policies that transfer to the real robot from a handful demonstrations and minimal reward function design; and (3) a comprehensive experimental validation and ablations demonstrating the effectiveness, scalability, and robustness of our method across diverse challenging dexterous tasks (Fig.~\ref{fig:tasks}).


\section{Related Work}
\label{sec:related-work}
\subsection{Imitation learning and teleoperation} 
Imitation learning enables autonomous robots by learning policies from expert demonstrations. Recent advances have focused on improving teleoperation systems to enhance the quality of expert data collection~\cite{wu2024gello, fu2024mobile, yang2024ace, si2024tilde, yin2025dexteritygen}. While these systems have significantly advanced robotic manipulation, they remain constrained by the inherent limitations of teleoperation, such as restricted operator feedback, system complexity, and latency. To mitigate some of these limitations, kinesthetic teaching offers an alternative by allowing operators to physically guide the robot in real-world demonstrations. Earlier studies in kinesthetic teaching primarily focused on controlling robotic arms and parallel jaw grippers~\cite{kronander2013learning, kormushev2011imitation}. More recently, \cite{chen2025dexforce} extended this approach to a multi-fingered robotic hand, without controlling the robot arm, by collecting force sensor data from the finger joints during demonstrations and applying diffusion policies~\cite{chi2023diffusionpolicy} that incorporate this force information. However, kinesthetics control of high-DoF robot systems-comprising both arms and hands-remains challenging, as operators often struggle to manipulate many degrees of freedom simultaneously. 

A more compelling strategy for collecting demonstrations involves directly capturing human interactions with the environment, thus avoiding the limitations introduced by using robots as intermediaries during data collection. Prior research has investigated learning policies from demonstrations collected in the wild using sensorized parallel jaw grippers~\cite{song2020grasping, shafiullah2023bringing, chi2024universal}. Other methods, such as those presented in~\cite{arunachalam2023dexterous, shaw2023videodex, sharma2019third}, have utilized human manipulation videos and applied motion re-targeting to robotic hands. More relevant to our work, \cite{wang2024dexcap} introduced a portable MoCap system for bi-manual dexterous robots, enabling direct real-world imitation learning. However, these techniques are often limited by the kinematic and dynamic mismatch between the operator hands and the robotic hands, requiring specialized learning algorithms to map the human hand movements to the robotic hands. Instead, we propose equipping demonstrators with a sensorized exoskeleton to collect human demonstrations, leveraging human sensory perception and motor skills while reducing the domain gap and improving data collection efficiency.

\subsection{Learning in simulation and sim-to-real transfer} 
Reinforcement learning (RL) in simulation has been widely adopted for learning generalizable manipulation skills for dexterous robotic hands, with domain randomization enabling the transfer of these skills to the real-world~\cite{andrychowicz2020learning,chen2022system,yin2023rotating,qi2023general,wang2024lessons}. However, training RL policies for highly dexterous tasks that can be transferred to the real world is non-trivial, often requiring excessive data and computational resources, careful reward function design, or an intricate learning scheme.

Previous research has shown that using demonstrations can facilitate RL training and improve learning efficiency~\cite{schaal1996learning, rajeswaran2017learning, vecerik2017leveraging, zhu2018reinforcement}. In this work, we also follow the demonstration-guided RL approach, where we use an auto-curriculum learning method, DemoStart~\cite{bauza2024demostart}, that leverages a small number of demonstrations and learns from sparse binary rewards. This approach, however, typically relies on dynamically feasible trajectories collected in simulation which can be challenging to obtain for high-DoF robot systems, such as robotic hands. Previous work has explored leveraging existing human hand motion datasets~\cite{rajeswaran2017learning, chen2024object} to guide RL training, but this still requires addressing the motion re-targeting problem resulting from the morphology gap between human and robotic hands. Another simulation-based approach is to directly synthesize hand motions to generate optimal trajectories by formulating it as a constrained optimization problem~\cite{ye2012synthesis, liu2008synthesis}. Similarly, we use trajectory optimization to convert collected real-world human demonstrations into dynamically feasible trajectories in simulation, making them suitable for training with auto-curriculum RL.
	

\section{Methods}
\label{sec:methods}
\begin{figure}[t]
\centering
\includegraphics[width=0.99\linewidth]{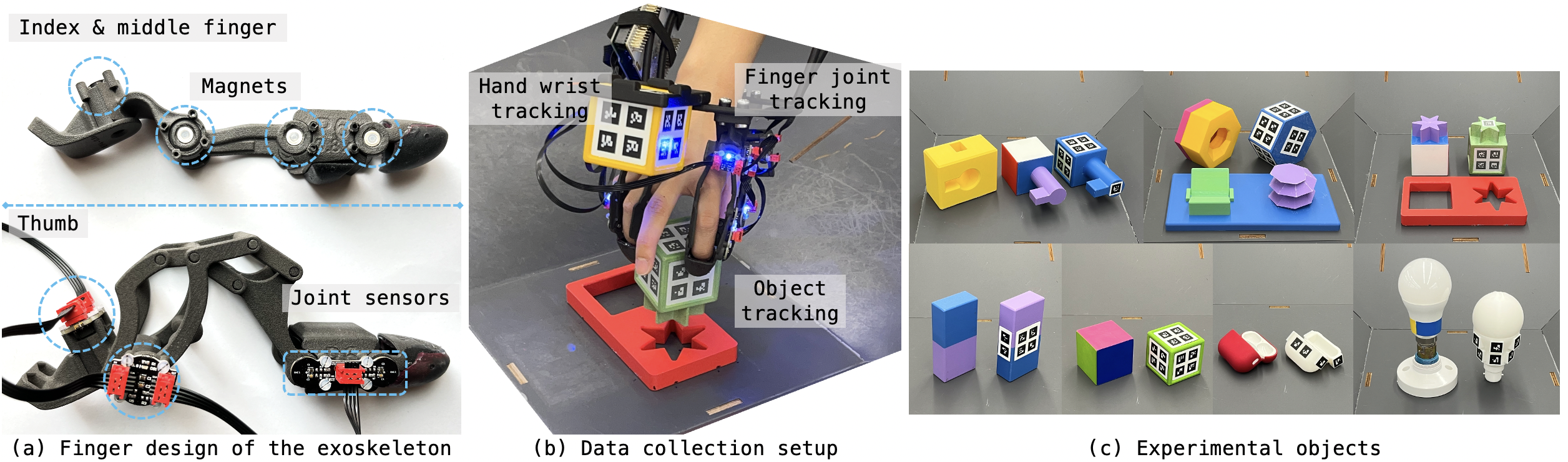}
\caption{(a) Finger design of the exoskeleton; (b) Sensorized data collection environment; (c) 3D-printed and real objects used for policy evaluation (w/o AR tags) and data collection (w/ AR tags).}
\label{fig:setup}
\end{figure}

Our proposed framework, \name, follows the real-to-sim-to-real pipeline illustrated in Fig.~\ref{fig:overview}. It begins by collecting direct human demonstrations via an exoskeleton, which are then processed through a simulation-based dynamics filter to recover dynamically feasible trajectories. Subsequently, an auto-curriculum RL method driven by simple binary rewards is bootstrapped using these trajectories and the learned policies are transferred from simulation to the real robot in a zero-shot manner.

\subsection{Direct human demonstration collection with an exoskeleton}
\label{subsec:methods_glove}
We propose the use of a sensorized wearable exoskeleton for data collection (Fig.~\ref{fig:setup} (b)), which allows operators to directly interact with the real environment. This device, in the form of a glove with rigid segments connected by joints, is designed to replicate the kinematic structure and joint limits of a robotic hand. In this work, we build an exoskeleton for the Shadow DEX-EE hand~\cite{dex}. The kinematic correspondence between the robotic hand and the exoskeleton allows for a direct one-to-one mapping from the human hand movements to the robotic hand, due to the shared kinematic constraints. The exoskeleton's mechanical structure (Fig.~\ref{fig:setup} (a)) is primarily assembled from low-cost 3D-printed components that accurately match the hand's palmar surface geometries. Soft elastomer pads on the fingertips and middle phalanxes provide tactile feedback to the operator during data collection. A joint position sensor embedded in each finger joint tracks the finger movements. In addition, we use a camera-based pose estimator to track and estimate the Cartesian poses of the exoskeleton and objects, noting that any other 6D pose trackers~\cite{guan2024survey} could be used as an alternative. Further details are provided in Appendix~\ref{sec:glove_appendix}.

\subsection{Dynamics filtering}
\label{sec:dynamics_filter}
Our framework requires dynamically feasible state trajectories in simulation to bootstrap policy learning. However, the collected raw real-world data is affected by hardware noise (including electronics and manufacturing errors) and visual tracking inaccuracies, resulting in trajectories that may not respect the simulated environment's dynamics. To clean them up, we use a sampling-based trajectory optimization method~\cite{howell2022predictive} as a dynamics filter to generate simulated trajectories that are dynamically feasible and closely track the recorded data.

For each time step, we first estimate the desired joint positions and object poses by interpolating the raw data collected in Sec.~\ref{subsec:methods_glove}. We then iteratively sample a set of control input sequences for a finite horizon, and select the sequence that minimizes a cost that measures the discrepancy between the real world measurements and the simulated state trajectories. Finally, we execute the first control input in this sequence, and repeat the same procedure for the rest of the demonstrated trajectory. 

The cost consists of a weighted sum of tracking terms on the 1) arm end-effector poses, 2) the object poses, 3) the arm joint positions, and 4) the distances between the fingertips and the vertices of the bounding box of the object being manipulated. We also add a penalty cost on the finger velocities as a regularization term. This task-independent solution encourages the filtered trajectories to match the demonstrated motions while respecting the simulated environment's dynamics. Further details are presented in Appendix~\ref{sec:mjpc_appendix}.

\subsection{Auto-curriculum reinforcement learning and distillation}
\label{subsec:rl}
We use DemoStart~\cite{bauza2024demostart}, an auto-curriculum RL method to train feature-based teacher policies in simulation and distill them to vision-based policies that transfer to the real robot. This method learns complex manipulation behaviors from only a sparse binary reward function and a handful of feasible state trajectories in simulation. The binary reward function is simple to define, relying solely on a small number of success criteria (as shown in Appendix~\ref{sec:tasks_appendix}): a reward of 1 is given if all criteria are met, and 0 otherwise. The state trajectories, obtained from Section~\ref{sec:dynamics_filter}, help guide the learning process by biasing the policy towards the human demonstrations. By leveraging RL, the resulting policies not only outperform the provided human demonstrations but also achieve greater robustness to a wider range of initial conditions.

Following~\cite{bauza2024demostart}, we train the teacher policies in a distributed actor-learner setup~\cite{espeholt2018impala} using privileged observations. We define an RL environment in simulation with a native initial state distribution for objects and robot poses ($S_{native}$) resembling those in the real robot environment, alongside the set of dynamically feasible states ($S_{demo}$) generated in Section~\ref{sec:dynamics_filter}. During training, the auto-curriculum RL method samples an initial state from either $S_{native}$ or $S_{demo}$, performs a handful of policy rollouts~(4 in our experiments) starting from the sampled state, and uses Zero-Variance Filtering\footnote{This consists of only selecting states where some of the rollouts failed and some succeed, in other words, they reach a mix of 0 and 1 final rewards, as introduced in~\cite{bauza2024demostart}.} to identify whether the state yields a strong learning signal. If so, then actors collect experience by executing the policy through episodes starting from this initial sampled state, and send data to the learner via a replay buffer~\cite{lin1992reinforcement}. Physical domain randomization is incorporated during training to improve policy robustness~\cite{andrychowicz2020learning}. 

The final stage involves distilling the trained teacher policies into vision-based student policies for zero-shot real-world deployment. This distillation process begins by rolling out episodes from the feature-based teacher policies in simulation to gather observation-action pairs. To mitigate sim-to-real gaps, domain randomization is applied during these rollouts. Subsequently, a student policy is trained via behavioral cloning using the collected simulated data and Action Chunking with Transformers (ACT)~\cite{zhao2023learning}, which we found to be more performant than the Perceiver-Actor-Critic~\cite{springenberg2024offline} implementation used in~\cite{bauza2024demostart} The resulting vision-based policy is then directly evaluated on the real robot. Further details are available in Appendix~\ref{sec:demostart_appendix}.

\section{Experimental Setup}
\label{sec:experimental_setup}
\begin{figure}[t]
\centering
\includegraphics[width=0.99\linewidth]{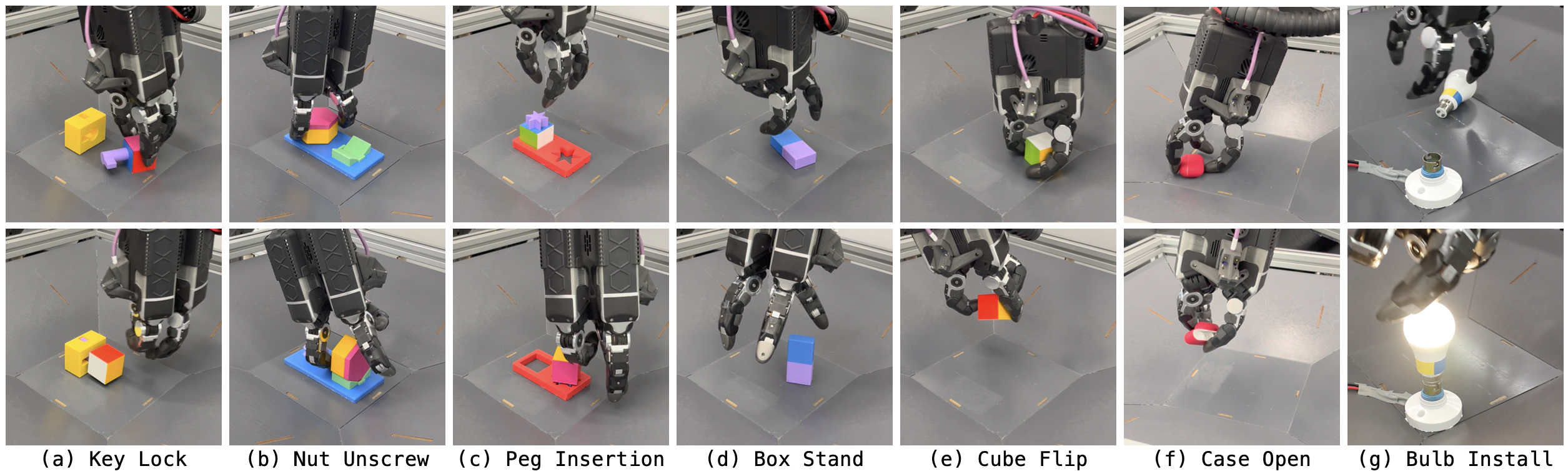}
\caption{The seven tasks used for experimental validation. See Appendix~\ref{sec:tasks_appendix} for task definitions.}
\label{fig:tasks}
\end{figure}

\subsection{Real-world and simulated setup}
Our real robot setup features a Kuka LBR iiwa 14 arm equipped with a {Shadow DEX-EE hand}~\cite{dex} as its end-effector. The arm is positioned beside a basket, with three fixed cameras pointing towards its center. The robotic hand has two cameras mounted on its wrist. We control the arm's end-effector Cartesian velocities (6 DoF) and the hand's joint positions (12 DoF). We replicate this setup in simulation using MuJoCo~\cite{todorov2012mujoco}. Both the simulation and real-world robot environments run at 20 Hz. Further details are in Appendix~\ref{sec:setup_appendix}.

To evaluate our proposed framework, we select seven representative and challenging dexterous manipulation tasks (Fig.~\ref{fig:tasks}): Key Lock, Nut Unscrew, Peg Insertion, Box Stand, Cube Flip, Case Open, and Bulb Install. These tasks demand a range of complex skills—such as grasping objects in random orientations, in-hand reorientation, articulated object manipulation, precise shape matching, and tight insertions—which are difficult to accomplish using traditional teleoperation methods. We believe their diversity and difficulty make them a compelling benchmark for assessing generalization, dexterity, and robustness in real-world robotic hand control.

Each experimental object has two versions: a target object and an AR-tagged replica (Fig.~\ref{fig:setup} (c)). The target objects consist of off-the-shelf items as well as custom 3D-printed objects. The AR-tagged objects are 3D-printed replicas of the target objects, with AR markers attached at pre-defined object surface locations to ease pose estimation during demonstrations. Note that the AR-tagged objects are only needed for data collection, as the target objects are always used during the real-world evaluation. We use the target object's model and textures during simulation and distillation to ensure that the policies can be transferred to the real world. 

\subsection{\namespace pipeline setup}
\paragraph{Demonstration collection}
During real-world data collection, we utilize the AR-tagged objects and the exoskeleton described in Section~\ref{subsec:methods_glove}. Each of the exoskeleton's finger joints has an embedded magnet and a Hall effect sensor, providing 12-dimensional joint angle data (three 4-DoF fingers). To estimate the arm's 6D end-effector pose, we attach a cube with 4×4 AR tags on each face to the exoskeleton's frame and use a Jacobian-based Inverse Kinematics solver~\cite{dmrobotics2021} to determine the arm's 7-DoF joint positions. To minimize visual occlusions during demonstrations, we employ a total of five basket cameras. The glove's sensor data is acquired and streamed at 200 Hz via a custom micro-controller unit board. The arm joint positions, end-effector poses, and object poses are recorded at 10 Hz and synchronized with the glove's finger joint data through ROS.

Table~\ref{tab:results} lists the number of demonstrations collected directly in the real world with the exoskeleton for each of the tasks, which ranges between 9 and 15. Note that the number of demonstrations collected is not indicative of the task difficulty, in fact in Section~\ref{sec:ablation-study} we show that \namespace can even work with just one demonstration. At the beginning of each episode, the operator wears the exoskeleton and starts with an open hand above the basket surface. All demonstrations are collected within a 15 seconds time limit, which shows the efficiency of our data collection. 

\paragraph{Dynamics filtering}
As described in Section~\ref{sec:dynamics_filter}, we use a sampling-based planner to track the real-world trajectories while respecting the simulated environments' dynamics. In our experiments, the cost terms are shared among tasks and the parameters for every task were initialized with the same values, with minimal tuning from one task to the next. We run the planner on each real-world demonstration in a loop for three hours to collect diverse successful trajectories for policy learning using the same reward function as in the RL environment. The number of successfully filtered trajectories for each task is shown in Table~\ref{tab:results} and ranges between 25 and 150. Further details are shown in Appendix~\ref{sec:mjpc_appendix}. 

\begin{table}[]
\centering
\begin{tabular}{lccccc}
\thickhline
 & \textbf{\# demos} &  \textbf{\# filtered} & \textbf{Teacher (Sim)} & \textbf{Student (Real)} & \textbf{Time limit (sec)} \\ 
 \hline
\textbf{Key Lock} & 13 & 68 & 99\% & 56\% & 15 \\
\textbf{Nut Unscrew} & 13 & 56 & 89\% & 50\% & 20 \\
\textbf{Peg Insertion} & 10 & 116 & 99\% & 54\%& 15  \\
\textbf{Box Stand} & 11 & 103 & 97\% & 94\% & 15 \\
\textbf{Cube Flip} & 9 & 132 & 95\% & 62\% & 15 \\
\textbf{Case Open} & 13 & 150 & 99\% & 56\% & 20 \\ 
\textbf{Bulb Install} & 15 & 25 & 98\% & 2\% & 20  \\
\thickhline
\end{tabular}
\caption{Experimental results. \textbf{\# of demos}: the number of demonstrations collected with the exoskeleton in the real world. \textbf{\# of filtered}: the number of demonstrations after applying dynamics filtering. \textbf{Teacher}: the success rate of the feature-based teacher policy during the episode roll-out in simulation (evaluated with $>$250,000 episodes); \textbf{Student}: the success rate of the vision-based student policy in the real environment (evaluated with 50 episodes); \textbf{Time limit}: the time limit for each episode during evaluation, in seconds.}
\label{tab:results}
\vspace{-3mm}
\end{table}

\paragraph{Policy learning and distillation}
\label{sec:policy-learning}
We feed the dynamically feasible simulated states into the DemoStart pipeline to train a feature-based teacher policy. For each task, we define a simple binary reward which is set to 1 if the task is achieved (Fig.~\ref{fig:tasks} illustrates examples of successful states), and zero otherwise. The task description and success criteria are included in Appendix~\ref{sec:tasks_appendix}. 

Then we distill the teacher policies into vision-based student polices for real-world robot deployment. 
The observations include the images from the three basket cameras and two wrist cameras, the arm joint positions, the arm end-effector poses, and the DEX hand finger joint positions. For each task, we roll out 250,000 episodes to generate trajectories of observation-action pairs in simulation, and use these to train an ACT policy which is deployed on the real-world robot zero-shot.


\section{Results and Discussion}
\label{sec:results}
Table~\ref{tab:results} presents the performance of our final vision-based student policies in solving the real-world tasks, as well as the performance of the feature-based teacher policies in simulation. 
Each task has a wide range of initial conditions, both in the simulated and real-world environments. More details of the task initial conditions are shown in Appendix~\ref{sec:tasks_appendix}. 
Consistent with DemoStart, our teacher policies achieved success rates of 95\% or higher for most tasks, except for 89\% for the Nut Unscrew task, evaluated over more than $250,000$ episodes in simulation. It is important to note that the teacher policy results reported in Table~\ref{tab:results} correspond to the specific policy used for distillation. We found that although we could obtain higher-performing teacher policies in simulation, this did not always translate to higher real-world performance. The lower success rate for the Nut Unscrew teacher policy can be attributed to the use of a refined simulation model, which improved sim-to-real transfer but slightly reduced performance in simulation. 

The distilled student policies demonstrated success rates exceeding 50\% across all real-world evaluations over 50 episodes, except for the Bulb Install task.
This represents a significant advancement over DemoStart, which only successfully transferred pick-and-place tasks that did not involve in-hand object re-orientation. This improvement highlights the advantage of leveraging expressive demonstrations to bootstrap reinforcement learning, enabling control policies to acquire dexterous behaviors from human demonstrations while the exploratory nature of RL enhances robustness for real-world deployment. The results also demonstrate the applicability of our method to solve highly challenging real-world tasks: opening an AirPods case in-hand, and inserting a light bulb into a bayonet connector with a tight fitting. 

At the time of writing, the Bulb Install task exhibits a non-zero but very low success rate (2\%). Despite this, we view this is as a positive result that showcases how expressive demonstrations can result in dexterous behaviors even in the presence of a large sim-to-real gap. In this task, we observed that MuJoCo’s soft-contact model led to object penetration, allowing the bulb to pass through the thin connector walls during the insertion phase in simulation. Consequently, the agent learned to exploit this discrepancy, explaining most real-world failures which happened during the insertion phase. Nonetheless, achieving zero-shot success in such a challenging real-world task highlights the value of high-quality demonstration data in guiding policy learning. We also note that improving the contact model of the simulator, while extremely relevant, is beyond the scope of this work.

\subsection{Ablation studies}
\label{sec:ablation-study}
\paragraph{Direct human demonstration vs teleoperation} 
We assess the benefits of direct object interaction by using the same exoskeleton described in Sec.~\ref{subsec:methods_glove} to teleoperate the robot in simulation. Teleoperation in simulation was chosen as it is required to generate dynamically feasible trajectories for our pipeline, effectively serving as a substitute for both real-world data collection and the dynamics filtering process. We limited our data collection to 10 minutes per episode and 1 hour per task, stopping at 1 hour or 10 successful episodes. Table~\ref{tab:data_collection} shows the success rates and average successful episode times for the Peg Insertion, Nut Unscrew, and Cube Flip tasks.

Unsurprisingly, direct demonstrations yielded higher success rates and lower data collection times. Teleoperation for the Peg Insertion task was six times slower due to the difficulty in flipping the peg in-hand, requiring a pick-and-place strategy. The Nut Unscrew task failed during teleoperation due to finger coordination challenges, with episodes timing out. The Cube Flip task was unsuccessful due to the inability of the user to perform in-hand rotation of the cube. We found that although teleoperation is a viable alternative for pick-and-place tasks, direct demonstrations are significantly more effective for dexterous manipulations involving in-hand reorientation and finger coordination, which we attribute to the force feedback that the operator gets during the interaction.

\begin{table}[]
\centering
\begin{tabular}{lcccccc}
\thickhline
 & \multicolumn{2}{ c }{\textbf{Peg Insertion}} & \multicolumn{2}{ c }{\textbf{Nut Unscrew}} & \multicolumn{2}{ c }{\textbf{Cube Flip}} \\ 
 \cline{2-3} \cline{4-5} \cline{6-7}
\textbf{} & \textbf{Teleop} & \textbf{Direct} & \textbf{Teleop} & \textbf{Direct} & \textbf{Teleop} & \textbf{Direct} \\ \hline
\textbf{\# of successful demos} & 9/10 & 10/10 & 4/10 & 10/10 & 0/10 & 10/10  \\
\textbf{Time (seconds)} & 48 & 8 & 393 & 15 & N/A & 10 \\ \thickhline
\end{tabular}
\caption{Number of successfully collected episodes with direct human demonstrations vs teleoperation, and the average duration of the successful episodes.}
\label{tab:data_collection}
\vspace{-5mm}
\end{table}

\paragraph{Directly using real-world raw data for RL policy learning}
Dynamics filtering provides two key benefits: 1) smoothing noisy raw sensor data into dynamically feasible states, and 2) augmenting the recorded data by generating more than one feasible trajectory for each demonstration. To assess the importance of using a dynamics filter, we directly fed the raw sensor data into the DemoStart pipeline to train policies for the Box Stand and Peg Insertion tasks. For the Box Stand task, learning failed to converge. For the Peg Insertion task, training took six times longer (18 million vs. 3 million policy updates). This demonstrates that it is possible to learn RL policies with the raw unfiltered data, but dynamics filtering greatly enhances the learning efficiency and robustness. Although in the future the dynamics filtering step could be merged into the auto-curriculum RL pipeline, we found that optimizing the tracking costs through RL was much harder than through trajectory optimization.

\paragraph{Data efficiency}
Our framework typically relies on 9 to 15 real-world demonstrations for sufficient data coverage. To evaluate the impact of using a smaller number of real-world demonstrations, we conducted an experiment with only one single real demonstration for solving the Key Lock task. We used the dynamically filtered trajectories from this single demonstration to initialize the policy training and applied our standard pipeline. While simulation performance of the teacher policy remained unchanged at 99\% success, real-world performance dropped to 17/50, compared to 28/50 with multiple demonstrations. Qualitatively, we observed that the learned policy was less robust especially during the grasping phase, causing the object to be dropped while trying to insert the key in the lock. This suggests that the lack of data diversity might negatively affect the robustness of policy's behavior and impact the sim-to-real transfer.


\section{Conclusions}
\label{sec:conclusion}

This work introduces \name, an efficient and scalable learning framework for dexterous manipulation with robotic hands. Our framework achieves over 50\% success on a wide range of tasks demonstrating challenging behaviors, including grasping objects in random configurations, in-hand re-orientation, articulated object manipulation, precise shape matching, and insertion. Through extensive ablation studies, we demonstrate that: 1) our exoskeleton demonstrations are more efficient and expressive compared to teleoperation, 2) dynamics filtering can improve the quality and quantity of the demonstrations, further improving learning efficiency; and 3) seeding RL with high-quality demonstrations improves generalizability, robustness, and enables complex dexterous behaviors on real robots.

\clearpage
\section{Limitations}
\label{sec:limitations}
\subsection{Sim-to-real gap}
This work focuses on developing a framework for collecting high-quality hand demonstrations and training policies that transfer to real robots in a zero-shot manner. A key limitation of this approach is the challenge of bridging the sim-to-real gap, which arises from inaccuracies in modeling the physical world in simulation. This issue is particularly critical in reinforcement learning, where policies may exploit simulation-specific artifacts, ultimately hindering real-world performance. As discussed in Section~\ref{sec:results}, this gap contributes to lower success rates in the Bulb Install task, where the insertion phase is not accurately captured in simulation. In recent years, significant advances in simulation fidelity and domain randomization techniques have helped reduce the sim-to-real gap, enabling more complex tasks to be realistically modeled. Looking forward, we expected that continued improvements in simulation will further expand the range of tasks that can be effectively trained and transferred to real-world settings.

\subsection{Motion re-targeting from human hands to robotic hands}
Utilizing an exoskeleton for data acquisition provided a direct one-to-one mapping and allowed the operator to leverage the robot's hand structure beyond just the fingertip positions. However, this approach may be difficult to generalize across different hand morphologies, requiring the design of an exoskeleton for each robotic hand. While 3D printing significantly accelerates our ability to create these exoskeletons, it may be impractical to do for a large number of hands. In the future, we could explore using generic data collection gloves~\cite{wang2024dexcap} or direct human hand motion capture~\cite{li2025maniptrans} to collect hand-object interaction data, and leverage dynamics filtering to re-target the motions to various robotic hands~\cite{lakshmipathy2024contactmpc}.

\subsection{Object pose tracking}
The current methodology relies on a set of cameras and 3D-printed, AR-tagged replicas for object pose estimation.  While this simplified the pose estimation problem for our setup, it introduced challenges for tracking small objects (e.g., an AirPods case) and restricted tracking to rigid bodies. Future work could investigate integrating state-of-the-art 6-DOF pose estimators~\cite{deng2019pose, wen2021bundletrack,wen2020se,wen2024foundationpose} to mitigate hardware dependencies, enhance tracking accuracy, and enable the system to track deformable and non-rigid objects.

\subsection{Deviation from demonstrated motions during policy learning}
In some cases, bootstrapping our RL method with the human demonstrations resulted in undesirable emergent behaviors. These included deviations from the demonstrated actions and physically unrealistic movements, such as exploiting task shortcuts and large joint velocities. To mitigate this, supplementary termination conditions were introduced to guide policy learning. Despite these terminations, certain deviations persisted, such as the policy learned to use two fingers to unscrew the nut although the operator used three fingers during the demonstrations. This presents the trade-off in constraining RL with demonstrations: we could force the policy's behaviors to match the demonstrations but this could lead to sub-optimal learning and increase the complexity of reward and termination criteria; alternatively, we could avoid defining additional restrictions, which would typically lead to undesired behaviors. Future work could focus on developing methods to constrain motion generation without compromising generalization capabilities during RL training.

\subsection{Mixing simulated and real-world data during distillation}
We applied extensive domain randomization and performed system identification to minimize the sim-to-real gap. While this approach enabled us to achieve over 50\% success rates on real-world tasks, there is considerable room for improvement. We explored incorporating real-world success episodes of our distilled policy evaluation back into a second ACT policy training round. We collected 100 successful real-world episodes on the Peg Insertion task, and tested two training strategies: 1) mixing our original simulated data and the newly collected real-world success data at a 9:1 ratio, and 2) using only real-world data. Real-world evaluations yielded 22/50 successes with the 9:1 mixed data and 32/50 with only the real-world data. Compared to 27/50 successes using only simulated data, real-to-real behavior cloning (BC) showed a slight improvement. We believe this can be used as a starting point to generate more data for online RL in the real world.

\subsection{Behavioral cloning on dynamically filtered data}
We assessed whether we could do behavioral cloning directly from the dynamically filtered data. Note that this is not possible to do directly from the real-world demonstration data as our method does not record any actions. To study this, we trained an ACT policy by directly using the dynamically filtered episodes of the Cube Flip task. The training dataset consisted of 586 successful episodes, augmented with visual randomization to ensure consistency with our standard pipeline. It is important to note that the control rate difference between the simulation used for MJPC~(200~Hz) and the real robot environment (20 Hz) was significant, which we attempted to address using zero-order action hold. This difference was due to a constraint on the open-sourced MJPC implementation~\cite{howell2022predictive}, which ties the control rate to the integration timestep of the simulation, that resulted in unstable simulations at lower rates. We found that the control rate mismatch introduced a substantial dynamics gap and resulted in zero success during evaluation, even within the simulation environment and the same initial conditions. Although other methods to address the control rate difference could be explored, such as averaging the actions or more advanced methods, we did not further explore this approach.

\acknowledgments{The authors would like to thank Iman Khan, Mohammed Umayir Ahmed, Neil Sreendra, Nicolo Pantano, Ahmed Bajaber, Alim Jalloh, Nathan Batchelor, Federico Casarini, Jingwei Zhang, Alexander Herzog, Joss Moore, Leonard Hasenclever, Serkan Cabi, Stefan Welker, Takuma Yoneda, and Yuval Tassa for their help with experiments and their advice.}


\bibliography{references}  

\clearpage


\section{Appendix}
\subsection{Method details}
\subsubsection{Data collection}
\label{sec:glove_appendix}
Our exoskeleton is designed with a glove form-factor that replicates the kinematic constraints of the Shadow DEX-EE (DEX) hand~\cite{dex}. Like the DEX hand, each finger possesses 4 DoF, and the fingers are arranged with one finger opposing the other two fingers. To ensure user comfort across varying hand sizes, the entire exoskeleton can be scaled. We employed a scaling factor of 0.8 in our experiments, which provided the most comfortable fit for the operator. Furthermore, to enhance thumb flexibility, we customized the glove's thumb with a virtual linkage structure, as illustrated in Fig.~\ref{fig:setup} (a). The exoskeleton's finger components are 3D-printed from rigid materials and assembled into a cohesive structure. Soft silicone elastomer pads cover the fingertips and middle phalanges, mirroring the DEX hand's finger surfaces and allowing users to experience haptic feedback through these padded contact points.

Finger movements are tracked by joint position sensors integrated into each joint of the exoskeleton. This is achieved by embedding a ring magnet within each joint, which moves relative to a mounted Hall-effect sensor. The resulting changes in the magnetic field are measured by the sensor to determine joint angles. These sensor readings are sequentially routed from the fingertips to the finger bases and then transmitted through a micro-controller unit board. To capture the hand's 6D pose, an AR tag cube is attached to the glove's wrist. Similarly, AR tags are affixed to experimental objects for pose tracking. Five cameras positioned around the workspace are used to calibrate and track these AR tags. All collected data—hand wrist pose, finger joint positions, and object poses—is synchronized at 10 Hz using ROS during data acquisition.

\subsubsection{Dynamics filtering with MJPC}
\label{sec:mjpc_appendix}
\begin{table}[h]
\centering
\begin{tabular}{lllll}
\thickhline
\textbf{Cost terms} & \textbf{Dimension} & \textbf{Norm Type} & \textbf{Cost weights} & \textbf{Norm parameters} \\ \hline
Hand TCP position & 3 & L22 & 10 & (0.02,2) \\
Hand TCP orientation & 3 & Quadratic & 100 & N/A\\
Object position & 3 & L22 & 25 & (0.02,2)\\ 
Object orientation & 3 & Quadratic & 100 & N/A\\ 
Arm joint position & 7 & Quadratic & 80 & N/A\\
Finger joint velocity & 12 & Quadratic & 0.01 & N/A\\
Fingertips$\{0|1|2\}$ to keypoints & 3 & L22 & 1 & (0.02,2)\\ \thickhline
\end{tabular}
\caption{Cost terms and parameters used for dynamics filtering. In our experiments, we use 8 keypoints per object - located at each of the vertices of the object's bounding box - and associate a cost between each keypoint and each of the fingertips. The L22 norm is part of the open-sourced MJPC implementation~\cite{howell2022predictive}, and is a generalization of the ``smooth-abs" function defined in~\cite{tassa2012norml22}.}.
\label{tab:mjpc_costs}
\end{table}

\begin{table}[h]
\centering
\begin{tabular}{ll}
\thickhline
\textbf{Planner configuration} & \textbf{Parameter} \\ \hline
Agent horizon & 0.25 \\
Agent timestep & 0.25 \\
Sampling trajectories & 40 \\
Sampling spline points & 3 \\
Sampling exploration & 0.08 \\\thickhline
\end{tabular}
\caption{Planner configurations and parameters for dynamics filtering.}
\label{tab:mjpc_params}
\end{table}

\begin{figure}[h]
\centering
\includegraphics[width=0.98\linewidth]{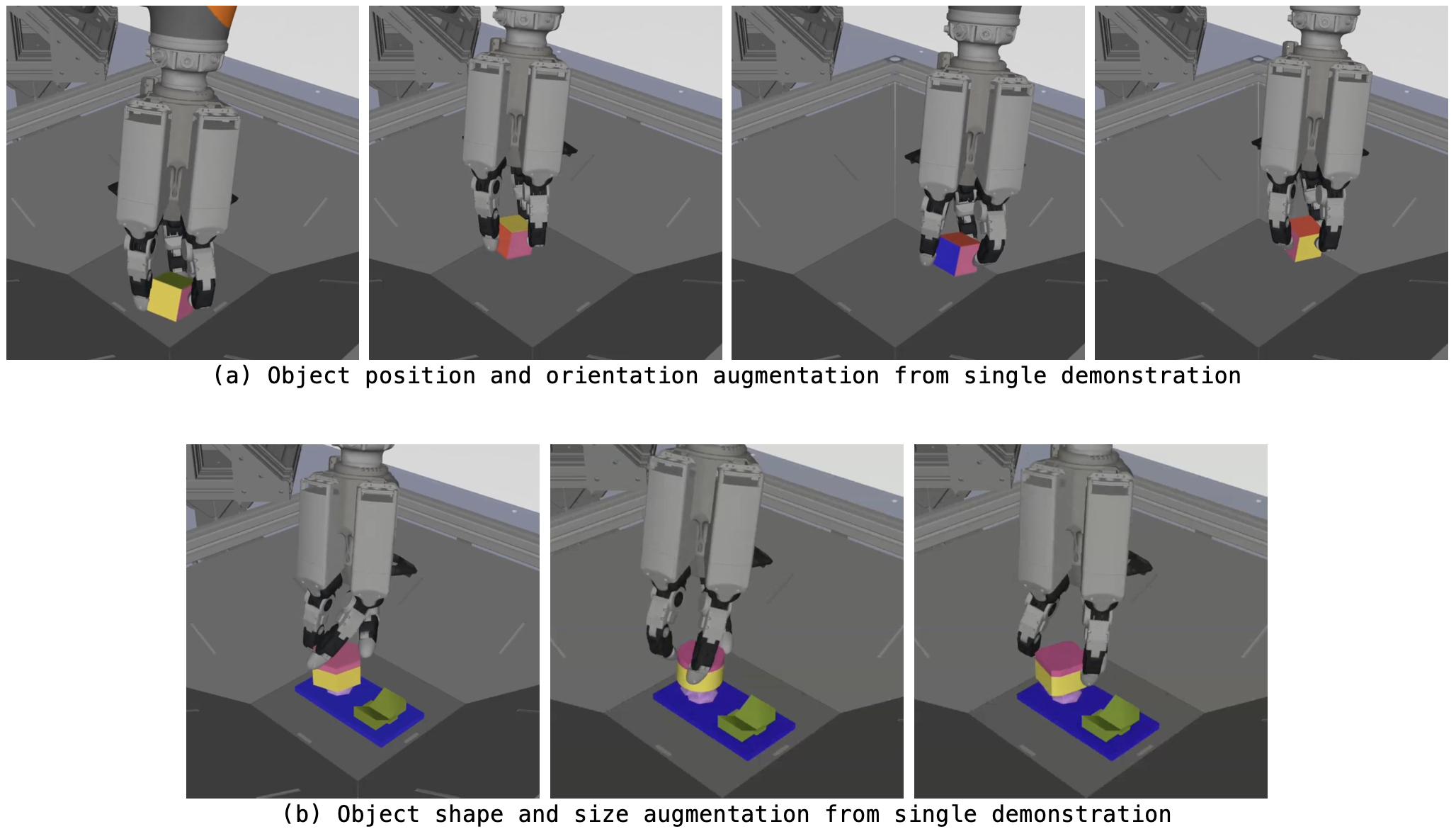}
\caption{We can augment the (a) object position and location, and (b) object shape and size from single demonstration with dynamics filtering.}
\label{fig:mjpc-augment}
\end{figure}

The collected real-world data often contains noise, leading to penetrations and jumps in the recorded motions. Using this data as demonstrations can lead to slower training times or negatively affect the success rate of our resultant policies. To address this, we employ sampling-based trajectory optimization with MJPC to recover dynamically feasible trajectories. We define cost terms to track the demonstrated motions, while ensuring that the trajectories respect the simulated environment's dynamics. The list of cost terms is shown Table~\ref{tab:mjpc_costs}.

For all tasks, the planner utilizes the cross entropy method with the parameters shown in Table~\ref{tab:mjpc_params}. In our experiments, the simulation state and control inputs are updated at a rate of 200~Hz. Since the glove's initial position during data collection usually differs from the RL training setup, we synthesize the first 20 frames by interpolating between the simulated environment's initial pose and the glove's initial pose. We also use the task's binary reward function as a success detector to filter out unsuccessful trajectories. The number of successful filtered episodes used for each task is shown Table~\ref{tab:results}.

We opt for a sampling-based planner due to its ability to tackle dexterous manipulation tasks with complex contact dynamics, which are challenging to optimize with gradient-based planners. Furthermore, the inherent randomness of the sampling approach yields a diverse set of trajectories even from a single demonstration. This diversity is beneficial for the auto-curriculum RL algorithm, as it provides a broader state distribution to reset the episodes in the RL environment.

Dynamics filtering serves two main purposes: recovering feasible trajectories and augmenting the demonstration data. For free floating objects like those in the Cube Flip and Box Stand tasks, we also augment the object poses by randomly sampling a constant pose offset that we apply to both the object and the robotic hand, as shown in Fig.~\ref{fig:mjpc-augment} (a). Similarly, we are also able to change the shape and size of the objects without the need to collect new demonstrations by replacing the simulated object, as shown in Fig.~\ref{fig:mjpc-augment} (b).

\subsubsection{Policy learning and distillation with DemoStart}
\label{sec:demostart_appendix}

\begin{table}[]
\centering
\begin{tabular}{ll}
\thickhline
\textbf{Feature-based policy observations} & \textbf{Dimension} \\ \hline
Arm joint position & 7 \\
Arm joint velocity & 7 \\
Arm end-effector position & 3 \\
Arm end-effector rotation matrix & 9 \\
Hand finger $\{0|1|2\}$ joint position & 12 \\
Hand finger $\{0|1|2\}$ joint velocity & 12 \\
Hand finger $\{0|1|2\}$ joint torques & 12 \\
Fingertip $\{0|1|2\}$ position & 3 \\
Fingertip $\{0|1|2\}$ rotation matrix & 9 \\ \thickhline
\end{tabular}
\caption{List of observations used by the feature-based teacher policy at each timestep. We stack the observations of the last three timesteps before sending it to the policy.}
\label{tab:demostart-obs}
\end{table}

\begin{table}[]
\centering
\begin{tabular}{ll}
\thickhline
\textbf{Vision-based policy observations} & \textbf{Dimension} \\ \hline
Arm joint position & 7 \\
Arm TCP position & 3 \\
Arm TCP rotation matrix & 9 \\
Hand finger $\{0|1|2\}$ joint position & 12 \\
Basket back left camera image & $222 \times 296 \times 3$ \\
Basket front left camera image & $222 \times 296 \times 3$ \\
Basket front right camera image & $222 \times 296 \times 3$ \\
Wrist front camera image & $222 \times 296 \times 3$ \\
Wrist back camera image & $222 \times 296 \times 3$\\ \thickhline
\end{tabular}
\caption{List of observations used by the vision-based student policy at each timestep.}
\label{tab:policy_params}
\end{table}

We formulate the policy learning problem as a Markov decision process (MDP), which we solve with RL. For each time step $t$, the agent predicts an action $a_{t} \in A$ based on its current state $s_t \in S$, receives a reward $r_{t+1} = R(s_t, a_t) \in R$, and transits to the next state $s_{t+1}$ given the transition probability distribution $p(\cdot | s_t, a_t)$.

Following the DemoStart~\cite{bauza2024demostart} pipeline, we use an auto curriculum learning method to train a feature-based teacher policy that we later distill into a vision-based policy. We first define two different state distributions, $S_{native}$ and $S_{demo}$. $S_{native}$ contains the set of initial states for the RL simulated environment that resemble the initial conditions of the real robot environment, usually supplemented through domain randomization techniques to ensure a good coverage of initial conditions. $S_{demo}$ is the set of dynamically feasible states obtained from the dynamics filter, which contains the states of every successful trajectory. As described in Section~\ref{subsec:rl}, DemoStart samples an initial state from either $S_{native}$ or $S_{demo}$ during training to reset the RL environment episodes, and identifies whether the state yields a strong learning signal. If it does, this initial state is used to collect experience by executing the policy through episodes starting from this state. All the episodes terminate after 10 seconds (200 environment steps) except for the Nut Unscrew task which terminates after 15 seconds (300 steps). The observations used during this training step is shown in Table~\ref{tab:demostart-obs}.

To reduce the sim-to-real domain gap, we randomize the physical properties during the teacher policy training, and both the physical and visual properties during distillation. The randomized physical parameters include the friction, mass, and inertia of objects and robot links, as well as the armature, damping, and frictionloss of the robot joints. The randomized visual parameters include the camera locations, lighting conditions, as well as the color and texture of every object. For every task, we also apply external force perturbations on the object after it has been grasped. These perturbations are applied for 10 timesteps and are either zero (with a 0.5 probability) or uniformly sampled from [0,1]~N, initiated only when at least two fingers are in contact with the object.

To train policies for real robot deployment, we distill the teacher feature-based policies into vision-based student policies using ACT~\cite{zhao2023learning}. Note that this differs from the original DemoStart pipeline which used a Perceiver-Actor-Critic~\cite{springenberg2024offline} for distillation, which we found was less performant than ACT for our tasks. During the policy training, we add additionally Gaussian noise ($std=0.3$) and photometric distortions including brightness $(-0.5,0.5)$, contrast $(0.5, 2.0)$, hue $(-0.5, 0.5)$, and saturation $(0.5, 1.5)$ to images. The observations used by the distilled vision-based student policies are shown in Table.~\ref{tab:policy_params}. During policy evaluation, we disable temporal aggregation and set the query frequency between one and four depending on the task.

\subsection{Experimental setup}
\label{sec:setup_appendix}

\begin{figure}[h]
\centering
\includegraphics[width=0.8\linewidth]{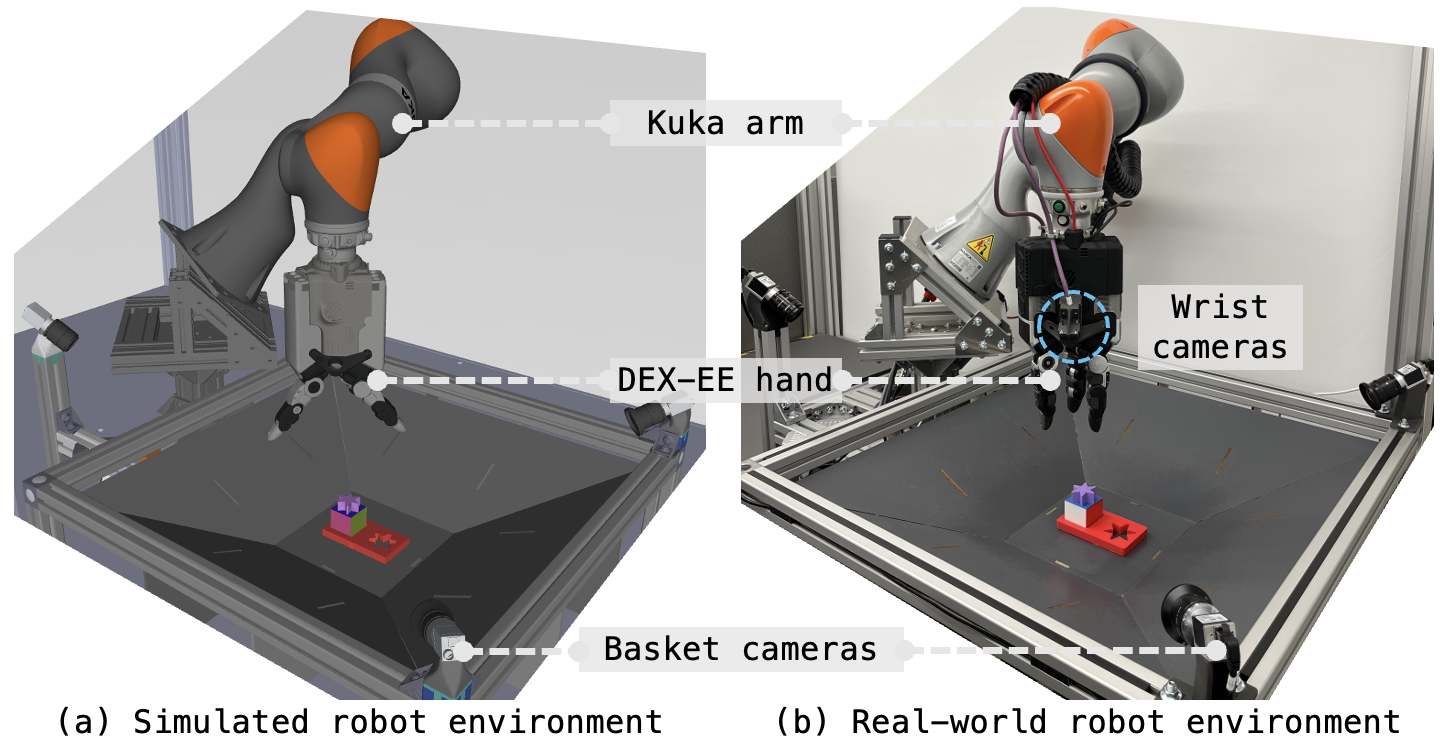}
\caption{Simulated and real-world robot environments.}
\label{fig:robot-env}
\end{figure}

The real-world and simulated setups are shown in Figure~\ref{fig:robot-env}. Our setup consists of a 7-DoF Kuka LBR iiwa 14 arm, a Shadow DEX-EE (DEX) hand~\cite{dex}, a basket with slanted walls, and five Basler ACA1920-40GC GigE cameras with standard lenses - three attached to the basket and two attached to the wrist of the DEX hand. The Kuka arm is controlled through a differential inverse kinematics controller~\cite{dmrobotics2021} that maps 6D Cartesian velocities to the joint velocities of the arm running at 200~Hz. The hand is controlled through a joint position controller running at 1~kHz. Both the simulation and real-world robot environments run at 20~Hz.

To collect real-world human demonstrations, we 3D-print object replicas with AR-tag holders and glue $13.1 \times 13.1$ mm x mm AR tags to them, as shown in Figure~\ref{fig:setup} (c). We collect the demonstrations with the same robot setup on which we run our real robot experiments. To reduce visual occlusion, we add two extra Basler ACA1920-40GC GigE cameras which we only use during the real-world data collection. To ensure consistent positions of the fixed objects, we use a jig to place the objects in pre-defined locations on the basket surface. During data collection, a human operator wears the exoskeleton, and demonstrates the tasks directly on the basket surface with the AR-tagged object replicas.

\subsection{Tasks}
\label{sec:tasks_appendix}

The seven tasks used for experimental validation are shown in Fig.~\ref{fig:tasks}. For each task, we present a high level description, as well as their success criteria, termination criteria, and initial conditions below:

\begin{enumerate}
    \item Key Lock: pick up the key, insert it into the lock, and turn it 90 degrees.
    \begin{itemize}
        \item Success criteria: a) the tip of the key is touching the back end of the lock; b) the key is turned 90 degrees; and c) the key is not being grasped.
        \item Termination criteria: a) the palm of the hand is not facing down, i.e. the angle between the normal of the transversal plane of the hand and the gravity vector is more than 30 degrees.
        \item Initial conditions in simulation: a) lock is fixed to a pre-defined location; b) key is initialized either in front of the lock or randomly around the basket surface, with a 0.95 and 0.05 probability, respectively.
        \item Initial conditions in real: a) lock is fixed to a pre-defined location; b) key is initialized in front of the lock.
    \end{itemize}
    \item Nut Unscrew: unscrew the nut by twisting 720 degrees, and place it onto the holder.
        \begin{itemize}
        \item Success criteria: a) the nut is standing on its side; b) the nut is on the holder; and 3) the nut is not being grasped.
        \item Termination criteria: a) the palm of the hand is not facing down, with a tolerance of 30 degrees.
        \item Initial conditions in simulation: a) bolt is fixed to a pre-defined location; b) nut is initialized either fully threaded, threaded at a random height, or randomly dropped in the basket with probability of 0.6, 0.3, 0.1, respectively.
        \item Initial conditions in real: a) bolt is fixed to a pre-defined location; b) nut is initialized fully threaded.
    \end{itemize}
    \item Peg Insertion: pick up the peg, re-orient the peg 180 degrees in-hand, and insert the peg into the star-shape socket.
        \begin{itemize}
        \item Success criteria: a) the peg is facing down; b) the tip center of the peg is touching the bottom center of the star socket; and c) the peg is not being grasped.
        \item Termination criteria:  a) the palm of the hand is not facing down, with a tolerance of 30 degrees; or 2) the prop orientation changes anywhere outside of a 3 x 3 cm area around the square socket. 
        \item Initial conditions in simulation: a) socket base is fixed to a pre-defined location; b) peg is either initialized in the square socket or randomly dropped in the basket, with a 0.8 and 0.2 probability, respectively.
        \item Initial conditions in real: a) socket base is fixed to a pre-defined location; b) peg is initialized in the square socket.
    \end{itemize}
    \item Box Stand: pick up the box, and make it stand vertically on the table with the blue segment on top.
        \begin{itemize}
        \item Success criteria: a) the box is standing with the blue segment on the top; and 2) the box is not being grasped.
        \item Termination criteria: a) the palm of the hand is not facing down, with a tolerance of 45 degrees.
        \item Initial conditions in simulation and real: box is randomly initialized on the basket surface.
    \end{itemize}
    \item Cube Flip: grasp the cube and re-orient it 180 degrees in-hand. 
        \begin{itemize}
        \item Success criteria: a) at least two fingers are in contact with the cube; b) cube is at least 5cm above the basket surface; and c) the cube has been rotated by 180 degrees from its initial orientation after being lifted.
        \item Termination criteria:  a) the palm of the hand is not facing down, with a tolerance of 30 degrees.
        \item Initial conditions in simulation and real: cube is randomly initialized on the basket surface.
    \end{itemize}
    \item Case Open: pick up the AirPods case, and open it in-hand.
        \begin{itemize}
        \item Success criteria: a) case is at least 5cm above the basket surface; and b) the case cap is open, i.e., the hinge joint position is at least 90 degrees.
        \item Termination criteria: the case is open (at least 10 degrees) while the case is in collision with the basket surface.
        \item Initial conditions in simulation and real: case is closed and randomly initialized on the basket surface.
    \end{itemize}
    \item Bulb Install: grasp the bulb, re-orient it in-hand, and install it into the bayonet socket. 
        \begin{itemize}
        \item Success criteria: 1) light bulb is upright; and 2) The two insertion pins of the bulb are in the two slots of the bayonet socket.
        \item Termination criteria: N/A.
        \item Initial conditions in simulation and real: a) bayonet socket is fixed to a pre-defined location; b) bulb is randomly initialized on the basket surface.
    \end{itemize}
\end{enumerate}

We use the same initial conditions for the objects during training and evaluation in simulation. To improve the robustness of the trained policy, we uniformly sample a position offset from [-1, 1] cm in the XY plane and an angle offset from [$-0.1$,$0.1$] radians along gravity vector, and apply these offsets to the initial pose of any fixed object. 

\end{document}